\documentclass[11pt,a4paper]{article}
\pdfoutput=1
\usepackage[hyperref]{naaclhlt2019}
\usepackage{times}
\usepackage{latexsym}
\usepackage{graphicx}
\usepackage{amsmath}
\usepackage[symbol]{footmisc}
\usepackage{booktabs}

\aclfinalcopy 

\title{A BERT Baseline for the Natural Questions}

\author{
  Chris Alberti\footnotemark[1] \;\;\;\; Kenton Lee \;\;\;\; Michael Collins\footnotemark[1] \\
  Google Research \\
  \small{\tt \{chrisalberti,kentonl,mjcollins\}@google.com}
}
\date{}

\begin{document}
\maketitle
\begin{abstract}
  This technical note describes a new baseline for the Natural Questions \cite{kwiatkowski2019nq}. Our model is based on BERT \cite{devlin2018bert} and reduces the gap between the model F1 scores reported in the original dataset paper and the human upper bound by 30\% and 50\% relative for the long and short answer tasks respectively. This baseline has been submitted to the official NQ leaderboard\footnote[2]{
  \url{https://ai.google.com/research/NaturalQuestions}
  }. Code, preprocessed data and pretrained model are available\footnote[3]{
  \url{https://github.com/google-research/language/tree/master/language/question_answering/bert_joint}}.
\end{abstract}

\footnotetext[1]{Also affiliated with Columbia University, work done at Google.}

\section{Introduction}

The release of BERT \cite{devlin2018bert} has substantially advanced the state-of-the-art in a number of NLP tasks, in question answering in particular. For example, as of this writing, the top 17 systems on the SQuAD 2.0 leaderboard \cite{rajpurkar2018know} and the top 5 systems on the CoQA leaderboard \cite{reddy2018coqa} are all based on BERT. The results obtained by BERT-based question answering models are also rapidly approaching the reported human performance for these datasets, with 2.5 F1 points of headroom left on SQuAD 2.0 and 6 F1 points on CoQA.

We hypothesize that the Natural Questions (NQ) \cite{kwiatkowski2019nq} might represent a substantially harder research challenge than question answering tasks like SQuAD 2.0 and CoQA, and that consequently NQ might currently be a good benchmark for the NLP community to focus on. The qualities that we think make NQ more challenging than other question answering datasets are the following: (1) the questions in NQ were formulated by people out of genuine curiosity or out of need for an answer to complete another task, (2) the questions were formulated by people before they had seen the document that might contain the answer, (3) the documents in which the answer is to be found are much longer than the documents used in some of the existing question answering challenges.

In this technical note we describe a BERT-based model for the Natural Questions. BERT performs very well on this dataset, reducing the gap between the model F1 scores reported in the original dataset paper and the human upper bound by 30\% and 50\% relative for the long and short answer tasks respectively. However, there is still ample room for improvement: 22.5 F1 points for the long answer task and 23 F1 points for the short answer task. 

The key insights in our approach are
\begin{enumerate}
    \item to jointly predict short and long answers in a single model rather than using a pipeline approach,
    \item to split each document into multiple training instances by using overlapping windows of tokens, like in the original BERT model for the SQuAD task,
    \item to aggressively downsample null instances (i.e. instances without an answer) at training time to create a balanced training set,
    \item to use the ``[CLS]'' token at training time to predict null instances and rank spans at inference time by the difference between the span score and the ``[CLS]'' score.
\end{enumerate}
We refer to our model as BERT\textsubscript{joint} to emphasize the fact that we are modeling short and long answers in a single model rather than in a pipeline of two models.

In the rest of this note we give further details on how the NQ dataset was preprocessed, we explain the modeling choices we made in our BERT-based model in order to adapt it to the NQ task, and we finally present our results.

\section{Data Preprocessing}
 
The Natural Questions (NQ) \cite{kwiatkowski2019nq} is a question answering dataset containing 307,373 training examples, 7,830 development examples, and 7,842 test examples. Each example is comprised of a google.com query and a corresponding Wikipedia page. Each Wikipedia page has a passage (or long answer) annotated on the page that answers the question and one or more short spans from the annotated passage containing the actual answer. The long and the short answer annotations can however be empty. If they are both empty, then there is no answer on the page at all. If the long answer annotation is non-empty, but the short answer annotation is empty, then the annotated passage answers the question but no explicit short answer could be found. Finally 1\% of the documents have a passage annotated with a short answer that is ``yes'' or ``no'', instead of a list of short spans.


Following \newcite{devlin2018bert} we tokenize every example in NQ using a 30,522 wordpiece vocabulary, then generate multiple instances per example by concatenating a ``[CLS]'' token, the tokenized question, a ``[SEP]'' token, tokens from the content of the document, and a final ``[SEP]'' token, limiting the total size of each instance to 512 tokens. For each document we generate all possible instances, by listing the document content starting at multiples of 128 tokens, effectively sliding a 512 token size window over the entire length of the document with a stride of 128 tokens. On average we generate 30 instances per NQ example. Each instance will be processed independently by BERT.


For each training instance we compute start and end token indices to represent the target answer span. If all annotated short spans are contained in the instance, we set the start and end target indices to point to the smallest span containing all the annotated short answer spans. If there are no annotated short spans but there is an annotated long answer span completely contained in the instance, we set the start and end target indices to point to the entire long answer span. If no short or long span can be found in the current instance, we set the target start and end indices to point to the ``[CLS]'' token. We dub the instances in the last category ``null instances''.

Given the large size of documents in NQ and the fact that 51\% of the documents are annotated as not having an answer to the query at all, we find that about 98\% of generated instances are null, therefore for training we downsample null instances by 50 times in order to obtain a training set that has roughly as many null instances as non-null instances. This leads to a training set that has approximately 500,000 instances of 512 tokens each.

We introduce special markup tokens in the document to give the model a notion of which part of the document it is reading. The special tokens we introduced are of the form ``[Paragraph=N]'', ``[Table=N]'', and ``[List=N]'' at the beginning of the N-th paragraph, list and table respectively in the document. This decision was based on the observation that the first few paragraphs and tables in the document are much more likely than the rest of the document to contain the annotated answer and so the model could benefit from knowing whether it is processing one of these passages. Special tokens are atomic, meaning that they are not split further by the wordpiece model.

We finally compute for each instance a target answer type as one of five values: ``short'' for instances that contain all annotated short spans, ``yes'' and ``no'' for yes/no annotations where the instance contains the long answer span, ``long'' when the instance contains the long answer span but there is no short or yes/no answer, and ``no-answer'' otherwise. Null instances correspond to the set of instances with the ``no-answer'' target answer type.

\begin{table*}
\begin{center}
\resizebox{\textwidth}{!}{%
 \begin{tabular}{r c c c c c c c c c c c c c} 
 \toprule
   & \multicolumn{3}{c}{Long Answer Dev}
   & \multicolumn{3}{c}{Long Answer Test} &
   & \multicolumn{3}{c}{Short Answer Dev}
   & \multicolumn{3}{c|}{Short Answer Test} \\
   &    P &    R &   F1 &    P &    R &   F1 &
   &    P &    R &   F1 &    P &    R &   F1 \\
 \midrule
   DocumentQA
   & 47.5 & 44.7 & 46.1 & 48.9 & 43.3 & 45.7 &
   & 38.6 & 33.2 & 35.7 & 40.6 & 31.0 & 35.1 \\
   DecAtt + DocReader
   & 52.7 & 57.0 & 54.8 & 54.3 & 55.7 & 55.0 &
   & 34.3 & 28.9 & 31.4 & 31.9 & 31.1 & 31.5 \\
   \textbf{BERT\textsubscript{joint} (this work)} 
   & \textbf{61.3} & \textbf{68.4} & \textbf{64.7} & \textbf{64.1} & \textbf{68.3} & \textbf{66.2} &
   & \textbf{59.5} & \textbf{47.3} & \textbf{52.7} & \textbf{63.8} & \textbf{44.0} & \textbf{52.1} \\
 \midrule
   Single Human
   & 80.4 & 67.6 & 73.4 & -    & -    & -    &
   & 63.4 & 52.6 & 57.5 & -    & -    & -    \\
   Super-annotator
   & 90.0 & 84.6 & 87.2 & -    & -    & -    &
   & 79.1 & 72.6 & 75.7 & -    & -    & -    \\
 \bottomrule
\end{tabular}}
\end{center}
\caption{Our results on NQ compared to the baselines in the original dataset paper and to the performance of a single human annotator and of an ensemble of human annotators. The systems used in previous NQ baselines are DocumentQA \cite{clark2017simple}, DecAtt \cite{parikh2016decomposable}, and Document Reader \cite{chen2017reading}.
}
\label{tab:results}
\end{table*}


\section{Model}

Formally, we define a training set instance as a four-tuple
   \[(c, s, e, t)\]
where $c$ is a context of 512 wordpiece ids (including question, document tokens and markup), $s, e \in \{0, 1, \ldots, 511\}$ are inclusive indices pointing to the start and end of the target answer span, and $t \in \{0, 1, 2, 3, 4\}$ is the annotated answer type, corresponding to the labels ``short'', ``long'', ``yes'', ``no'', and ``no-answer''.

We define the loss of our model for a training instance to be
\begin{align*}
    L &= - \log p(s, e, t | c) \\
      &= - \log p_{\mbox{start}}(s | c) - \log p_{\mbox{end}}(e | c) \\
      &\;\;\;\; - \log p_{\mbox{type}}(t | c),
\end{align*}
where each probability $p$ is obtained as a softmax over scores computed by the BERT model as follows:
\begin{align*}
    p_{\mbox{start}}(s | c) &= \frac{\exp(f_{\mbox{start}}(s, c; \theta))}
    {\sum_{s'}{\exp(f_{\mbox{start}}(s', c; \theta))}}, \\
    p_{\mbox{end}}(e | c) &= \frac{\exp(f_{\mbox{end}}(e, c; \theta))}
    {\sum_{e'}{\exp(f_{\mbox{end}}(e', c; \theta))}}, \\
    p_{\mbox{type}}(t | c) &= \frac{\exp(f_{\mbox{type}}(t, c; \theta))}
    {\sum_{t'}{\exp(f_{\mbox{type}}(t', c; \theta))}},
\end{align*}
where $\theta$ represents the BERT model parameters and $f_{\mbox{start}}$, $f_{\mbox{end}}$, $f_{\mbox{type}}$ represent three different outputs derived from the last layer of BERT.

At inference time we score all the contexts from each document and then rank all document spans $(s, e)$ by the score
\begin{align*}
   g(c, s, e) &= f_{\mbox{start}}(s, c; \theta) \\
        &+ f_{\mbox{end}}(e, c; \theta) \\
        &- f_{\mbox{start}}(s=\mbox{[CLS]}, c; \theta) \\
        &- f_{\mbox{end}}(e=\mbox{[CLS]}, c; \theta)
\end{align*}
and return the highest scoring span in the document as the predicted short answer span. Note that $g(c, s, e)$ is exactly the log-odds between the likelihood of an answer span (defined by the product $p_{\mbox{start}} \cdot p_{\mbox{end}}$) and the ``[CLS]'' span.

We select the predicted long answer span as the DOM tree top level node containing the predicted short answer span, and assign to both long and short prediction the same score equal to the maximum value of $g(c, s, e)$ for the document.

We opted to limit the complexity of this baseline model by always outputting a single short answer as prediction and we rely on the official NQ evaluation script to set thresholds to decide which of our predictions should be changed to having only a long answer or no answer at all. We expect that improvements can be obtained by combining start/end and answer type outputs to sometimes predict yes/no answers instead of always predicting a span as the short answer. We also expect additional improvements to be achievable by extending the model to be able to emit short answers comprised of multiple disjoint spans.

\section{Experiments}

We initialized our model from a BERT model already finetuned on SQuAD 1.1 \cite{rajpurkar2016squad}. We then further finetuned the model on the training instances precomputed as described in Section 2. We trained the model by minimizing loss $L$ from Section 3 with the Adam optimizer \cite{kingma2014adam} with a batch size of 8. As is common practice for BERT models, we only tuned the number of epochs and the initial learning rate for finetuning and found that training for 1 epoch with an initial learning rate of $3 \cdot 10^{-5}$ was the best setting.

Evaluation completed in about 5 hours on the NQ dev and test set with a single Tesla P100 GPU. 

The results obtained by our model are shown in Table \ref{tab:results}. Our BERT model for NQ performs dramatically better than the models presented in the original NQ paper. Our model closes the gap between the F1 score achieved by the original baseline systems and the super-annotator upper bound by 30\% for the long answer NQ task and by 50\% for the short answer NQ task. However NQ appears to be still far from being solved, with more than 20 F1 points of headroom for both the long and short answer tasks.

\section{Conclusion}
We presented a BERT-based model \cite{devlin2018bert} as a new baseline for the newly released Natural Questions \cite{kwiatkowski2019nq}.

We hope that this baseline can constitute a good starting point for researchers wanting to create better models for the Natural Questions and for other question answering datasets with similar characteristics.

\section{Acknowledgements}
We would like to thank Ankur Parikh, Daniel Andor, Emily Pitler, Jacob Devlin, Kristina Toutanova, Ming-Wei Chang, Slav Petrov, Tom Kwiatkowski and the entire Google AI Language team for many valuable suggestions and help in carrying out this work.

\bibliography{main}
\bibliographystyle{acl_natbib}

\end{document}